\pdfoutput=1

\documentclass[11pt]{article}

\usepackage[]{acl}

\usepackage{times}
\usepackage{latexsym}

\usepackage[T1]{fontenc}

\usepackage[utf8]{inputenc}

\usepackage{microtype}

\usepackage{trackchanges}

\usepackage{amsmath,amsfonts,amssymb,bm}
\usepackage{pifont} 
\usepackage{graphicx,epsfig,fancybox} 
\usepackage{float}
\usepackage{color} 
\usepackage[ruled,vlined]{algorithm2e} %
\usepackage{algorithmic}
\usepackage{multirow}
\usepackage{comment}
\usepackage{hyperref}
\usepackage{natbib} 
\usepackage{subfigure}

\renewcommand{\vec}[1]{\boldsymbol{#1}}

\newcommand{\mat}[1]{\mathbf{#1}}

\newcommand{\vth}[0]{\vec{\theta}}
\newcommand{\vphi}[0]{\vec{\phi}}

\usepackage{adjustbox}
\usepackage{array}
\usepackage{booktabs}

\newcolumntype{R}[2]{%
    >{\adjustbox{angle=#1,lap=\width-(#2)}\bgroup}%
    l%
    <{\egroup}%
}

\addeditor{YJ}
\addeditor{WD}
\addeditor{LW}
\addeditor{JZ}
\usepackage{wrapfig}
\newcommand{\newparagraph}[1]{\noindent\textbf{#1}}

\usepackage{tikz}
\usetikzlibrary{shapes.geometric, positioning}
\usetikzlibrary{arrows,shapes,calc}

\tikzstyle{randomvar}=[circle,
thick,
minimum size=1.2cm,
draw=black]

\tikzstyle{hidden}=[circle,
double,
double distance=1pt,
thick,
minimum size=1.2cm,
draw=black]

\tikzstyle{obs}=[circle,
thick,
minimum size=1.2cm,
draw=black!80,
fill=black!20]

\title{Diverse Text Generation via Variational Encoder-Decoder Models with Gaussian Process Priors}

\author{Wanyu Du$^1$\ , Jianqiao Zhao$^2$\ , Liwei Wang$^2$\ , Yangfeng Ji$^1$\\
$^1$Department of Computer Science, University of Virginia \\
$^2$Department of Computer Science and Engineering, The Chinese University of Hong Kong \\
\texttt{\{wd5jq,yangfeng\}@virginia.edu} \\
\texttt{\{jqzhao,lwwang\}@cse.cuhk.edu.hk} \\
}

\begin{document}
\maketitle

\begin{abstract}
  Generating high quality texts with high diversity is important for many NLG applications, but current methods mostly focus on building deterministic models to generate higher quality texts and do not provide many options for promoting diversity.
  In this work, we present a novel latent structured variable model to generate high quality texts by enriching contextual representation learning of encoder-decoder models.
  Specifically, we introduce a stochastic function to map deterministic encoder hidden states into random context variables. 
  The proposed stochastic function is sampled from a Gaussian process prior to (1) provide infinite number of joint Gaussian distributions of random context variables (diversity-promoting) and (2) explicitly model dependency between context variables (accurate-encoding). 
  To address the learning challenge of Gaussian processes, we propose an efficient variational inference approach to approximate the posterior distribution of random context variables.
  We evaluate our method in two typical text generation tasks: paraphrase generation and text style transfer.
  Experimental results on benchmark datasets demonstrate that our method improves the generation quality and diversity compared with other baselines.
  \footnote{Code and data are available at \url{https://github.com/wyu-du/GP-VAE}.}
\end{abstract}

\section{Introduction}
\label{sec:intro}

Generating high quality texts with high diversity is an important requirement for many text generation applications, such as paraphrase generation \citep{prakash2016neural,li-etal-2018-paraphrase}, style transfer \citep{jhamtani2017shakespearizing,rao-tetreault-2018-dear}, dialog generation \citep{sordoni2015hierarchical,serban2016building}, etc.
The encoder-decoder framework \citep{cho-etal-2014-learning,sutskever2014sequence} is widely adopted \citep{bahdanau2014neural,luong2015effective,gu-etal-2016-incorporating, see-etal-2017-get,DBLP:journals/corr/abs-1910-10683} to generate high-quality texts, where an encoder is applied to learn contextual information from source texts and a decoder is used to generate texts for target tasks.
To improve the diversity of generated texts, prior works propose to introduce variations into either the encoder \citep{bahuleyan2018variational,deng2018latent,ijcai2019-727,cho-etal-2019-mixture,qian-cheung-2019-enhancing,wu-etal-2020-encoder,duan-etal-2020-pre,sun-etal-2021-generating} or the decoder \citep{vijayakumar2016diverse,holtzman2019curious,he2018sequence,shen2019mixture}.
However, it is difficult to incorporate meaningful variations into encoder-decoder models without hurting the quality of generated texts.

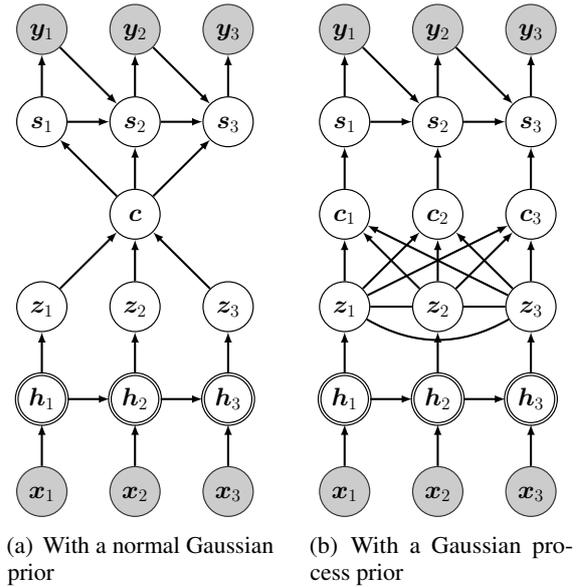
\begin{figure}[t]
  \centering
  \subfigure[With a normal Gaussian prior]{
    \scalebox{0.55}{\begin{tikzpicture}[>=latex,text height=2ex,text depth=0.25ex,line width=0.5mm]
  \tikzstyle{every node}=[font=\LARGE]
  \node[hidden] (h1) {$\vec{h}_1$};
  \node[hidden, right=of h1] (h2) {$\vec{h}_2$};
  \node[hidden, right=of h2] (h3) {$\vec{h}_3$};
  \node[randomvar, above=of h1] (z1) {$\vec{z}_1$};
  \node[randomvar, above=of h2] (z2) {$\vec{z}_2$};
  \node[randomvar, above=of h3] (z3) {$\vec{z}_3$};
  \node[obs, below=of h1] (x1) {$\vec{x}_1$};
  \node[obs, below=of h2] (x2) {$\vec{x}_2$};
  \node[obs, below=of h3] (x3) {$\vec{x}_3$};
  \node[randomvar, above=of z2] (c) {$\vec{c}$};
  \node[randomvar, above=of c] (s2) {$\vec{s}_2$};
   \node[randomvar, left=of s2] (s1) {$\vec{s}_1$};
  \node[randomvar, right=of s2] (s3) {$\vec{s}_3$};
  \node[obs, above=of s1] (y1) {$\vec{y}_1$};
  \node[obs, above=of s2] (y2) {$\vec{y}_2$};
  \node[obs, above=of s3] (y3) {$\vec{y}_3$};
  \draw[->] (z1) -- (c);
  \draw[->] (z2) -- (c);
  \draw[->] (z3) -- (c);
  \draw[->] (s1) -- (s2);
  \draw[->] (s1) -- (y1);
  \draw[->] (y1) -- (s2);
  \draw[->] (s2) -- (s3);
  \draw[->] (s2) -- (y2);
  \draw[->] (y2) -- (s3);
  \draw[->] (s3) -- (y3);
  \draw[->] (c) -- (s1);
  \draw[->] (c) -- (s2);
  \draw[->] (c) -- (s3);
  \draw[->] (h1) -- (h2);
  \draw[->] (h2) -- (h3);
  \draw[->] (x1) -- (h1);
  \draw[->] (x2) -- (h2);
  \draw[->] (x3) -- (h3);
  \draw[->] (h1) -- (z1);
  \draw[->] (h2) -- (z2);
  \draw[->] (h3) -- (z3);
\end{tikzpicture}}
    \label{fig:vae-normal}
  }
  \quad
  \subfigure[With a Gaussian process prior]{
    \scalebox{0.55}{\begin{tikzpicture}[>=latex,text height=2ex,text depth=0.25ex,line width=0.5mm]
  \tikzstyle{every node}=[font=\LARGE]
  \node[hidden] (h1) {$\vec{h}_1$};
  \node[hidden, right=of h1] (h2) {$\vec{h}_2$};
  \node[hidden, right=of h2] (h3) {$\vec{h}_3$};
  \node[randomvar, above=of h1] (z1) {$\vec{z}_1$};
  \node[randomvar, above=of h2] (z2) {$\vec{z}_2$};
  \node[randomvar, above=of h3] (z3) {$\vec{z}_3$};
  \node[obs, below=of h1] (x1) {$\vec{x}_1$};
  \node[obs, below=of h2] (x2) {$\vec{x}_2$};
  \node[obs, below=of h3] (x3) {$\vec{x}_3$};
  \node[randomvar, above=of z1] (c1) {$\vec{c}_1$};
  \node[randomvar, above=of z2] (c2) {$\vec{c}_2$};
  \node[randomvar, above=of z3] (c3) {$\vec{c}_3$};
  \node[randomvar, above=of c1] (s1) {$\vec{s}_1$};
  \node[randomvar, above=of c2] (s2) {$\vec{s}_2$};
  \node[randomvar, above=of c3] (s3) {$\vec{s}_3$};
  \node[obs, above=of s1] (y1) {$\vec{y}_1$};
  \node[obs, above=of s2] (y2) {$\vec{y}_2$};
  \node[obs, above=of s3] (y3) {$\vec{y}_3$};
  \draw[->] (c1) -- (s1);
  \draw[->] (c2) -- (s2);
  \draw[->] (c3) -- (s3);
  \draw[->] (z1) -- (c1);
  \draw[->] (z1) -- (c2);
  \draw[->] (z1) -- (c3);
  \draw[->] (z2) -- (c1);
  \draw[->] (z2) -- (c2);
  \draw[->] (z2) -- (c3);
  \draw[->] (z3) -- (c1);
  \draw[->] (z3) -- (c2);
  \draw[->] (z3) -- (c3);
  \draw[->] (s1) -- (s2);
  \draw[->] (s1) -- (y1);
  \draw[->] (y1) -- (s2);
  \draw[->] (s2) -- (s3);
  \draw[->] (s2) -- (y2);
  \draw[->] (y2) -- (s3);
  \draw[->] (s3) -- (y3);
  \draw[->] (h1) -- (h2);
  \draw[->] (h2) -- (h3);
  \draw[->] (x1) -- (h1);
  \draw[->] (x2) -- (h2);
  \draw[->] (x3) -- (h3);
  \draw[->] (h1) -- (z1);
  \draw[->] (h2) -- (z2);
  \draw[->] (h3) -- (z3);
  \draw (z1) -- (z2);
  \draw (z2) -- (z3);
  \draw (z1) [out=-30,in=-150] to (z3);
\end{tikzpicture}}
    \label{fig:vae-gp}
  }
  \caption{A simple illustration on a variational encoder-decoder model with (a) an normal Gaussian prior and (b) a Gaussian process prior. 
  With a normal Gaussian prior, each hidden state $\vec{h}_i$ will be mapped into a random vector $\vec{z}_i$ independently; while a Guassian process prior imposes dependency constraints among $\{\vec{z}_i\}$. Double circles on $\{\vec{h}_i\}$ indicate they are deterministic variables.}
\end{figure}

Promoting diversity at the encoder side mainly focuses on modelling the probabilistic distribution of contextual representations.
Some prior works \citep{deng2018latent,bahuleyan2018variational} propose to model attention alignments between encoder and decoder hidden states as latent variables, and generate diverse texts by sampling from the latent attention variables. 
Other existing works \citep{ijcai2019-727,liu2019transformer,shinoda-etal-2021-improving,sun-etal-2021-generating} directly apply conditional variational autoencoders to model encoder hidden states as latent variables, and generate high-diversity texts by sampling from the latent context variables. 
However, when modelling the latent variables, they treat each latent variable as independent to each other, which inevitably causes the loss of some contextual information during learning, as shown in Figure \autoref{fig:vae-normal}.

Other works turns towards designing diversity-promoting decoding strategies at the decoder side, such as diverse beam search \citep{vijayakumar2016diverse}, top-k sampling \citep{fan-etal-2018-hierarchical}, and nucleus sampling \citep{holtzman2019curious}.
But for those decoding strategies, there is often a trade-off between quality and diversity, and the generation models have to sacrifice quality for a higher diversity.
Another line of works suggest to learn a mixture of expert encoders \citep{cho-etal-2019-mixture} or decoders \citep{he2018sequence,shen2019mixture}, and generate diverse texts by sampling from different encoders or decoders.
While different expert encoders or decoders can introduce some diversity, the model capacities are limited within the pre-defined set of experts.

In this work, we propose a novel approach to introduce context-aware variations into the encoder in order to generate high-quality and high-diversity texts.
For an encoder-decoder model, we introduce a stochastic function to map deterministic encoder hidden states $\{\vec{h}_i\}$ into a set of {random context variables} $\{\vec{z}_i\}$. 
The advantage of this stochastic function is that it explicitly models the dependency between each context variable, as shown in Figure \autoref{fig:vae-gp}, which can help preserve more semantic information from source texts.
During generation, the decoder generates diverse outputs conditioning on sampled different context variables.
In other words, by learning a stochastic function on top of \emph{one} deterministic encoder, the proposed approach offers \emph{many} versions of random context variables for a decoder to generate diverse texts. 

\begin{figure*}[t]
  \centering
  \subfigure[Variational encoder-decoder model with a normal Gaussian prior. Note that we simplify the latent variable $\vec{z}_i$ from a vector to a scalar in order to plot out the Gaussian distribution for better illustration.]{
    \includegraphics[width=0.9\linewidth]{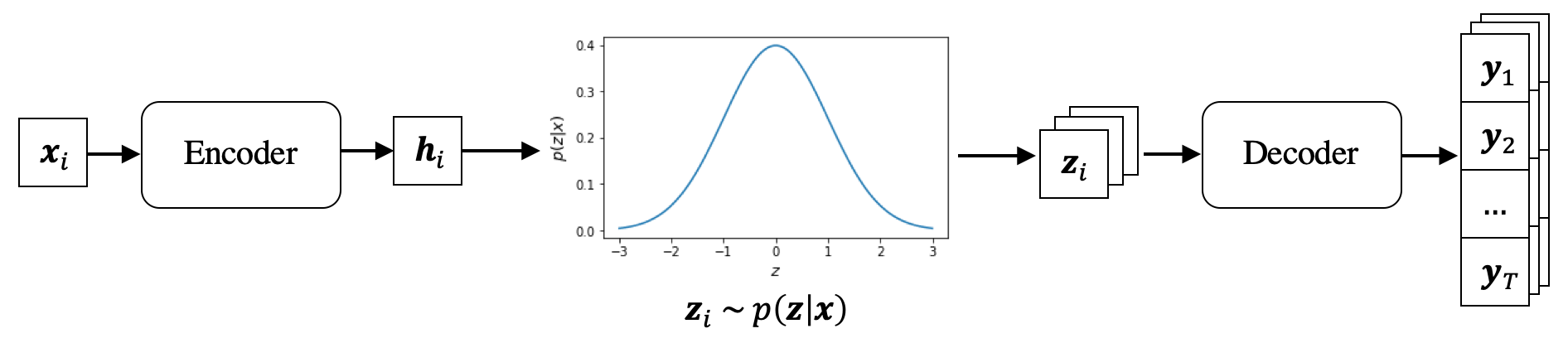}
    \label{fig:seq2seq-normal}
  }
  \quad
  \subfigure[Variational encoder-decoder model with a GP prior. Note that we simplify the latent variable $\vec{z}_i$ from a vector to a scalar in order to plot out the joint Gaussian distribution for better illustration.]{
    \includegraphics[width=0.9\linewidth]{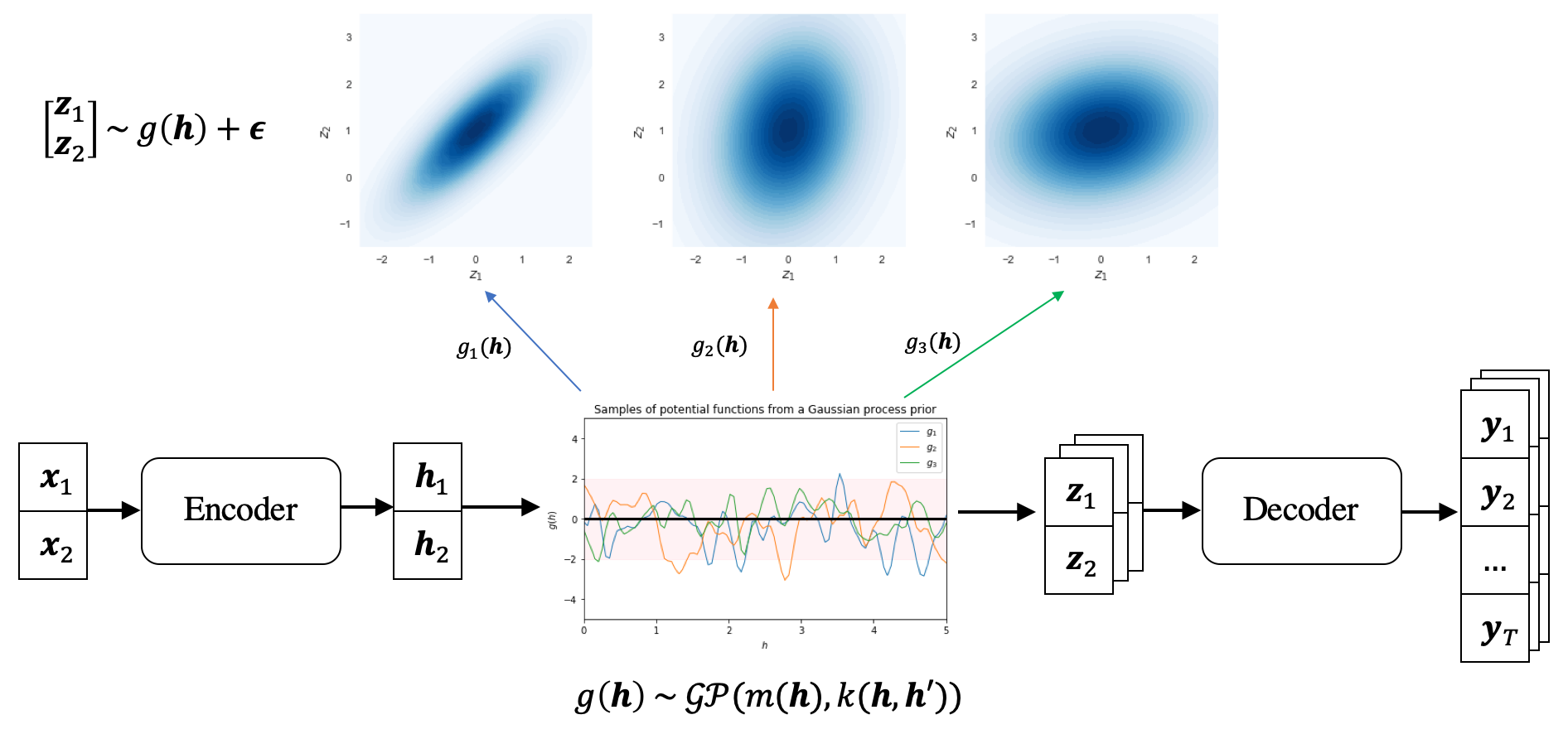}
    \label{fig:seq2seq-gp}
  }
  \caption{A simple illustration for comparison between our GP priors and the priors in conditional variational autoencoders \cite{NIPS2015_8d55a249} under the variational encoder-decoder framework. }
\end{figure*}

To learn the stochastic function over hidden states, we propose a Gaussian process prior \citep[GP]{rasmussen2006gaussian} to model the joint distribution of all encoder hidden states.
The major differences between GP priors and other priors used in previous works \citep{bahuleyan2018variational,deng2018latent,ijcai2019-727,cho-etal-2019-mixture,wu-etal-2020-encoder,duan-etal-2020-pre,shinoda-etal-2021-improving,sun-etal-2021-generating} have two-folds: 
(1) GP priors explicitly model the dependency between latent variables of varying sizes as illustrated in Figure \autoref{fig:vae-gp}, while previous works consider latent variables as independent with each other as shown in Figure \autoref{fig:vae-normal};
(2) GP priors provide \emph{infinite} number of joint Gaussian distributions of latent variables as shown in Figure \autoref{fig:seq2seq-gp}, while previous works have to pre-define a fixed set of Gaussian distributions (e.g. a standard normal distribution, or a mixture of Gaussian distributions) with the risk of experiencing the posterior collapse problem \citep{bowman-etal-2016-generating,pmlr-v80-kim18e,pmlr-v89-dieng19a}. 
Besides, the proposed random function only introduces variations into the encoder, and is orthogonal to diversity-promoting decoding strategies at the decoder side. 
Users can freely adopt different decoding strategies to further encourage diverse generation outputs.

The major contributions of this work are three-fold:
\begin{enumerate}
    \item We propose a novel method to introduce context-aware variations into encoder-decoder models, which can help the model learn rich contextual representations and also promote diversity in generation.
    \item We propose an efficient variational inference method to approximate the joint distribution of fully-connected random context variables.
    \item We test our proposed method in both LSTM-based \citep{{see-etal-2017-get}} and Transformers-based \citep{DBLP:journals/corr/abs-1910-10683} encoder-decoder models on paraphrase generation and style transfer tasks. Empirical experimental results show that, on one hand, the proposed method can generate higher quality texts than deterministic encoder-decoder models and conditional variational auto-encoders; on the other hand, it also supports diverse generation by conditioning on different sets of sampled random context variables.
\end{enumerate}

\section{Model Description}
This section discusses our novel latent structured variable model on learning rich context representations by transforming the hidden states from a deterministic encoder into random hidden states via stochastic functions.

\subsection{Encoding with Stochastic Functions}
\label{subsec:generative}

Let $\vec{x}_{1:N}=\{\vec{x}_i\}_{i=1}^N$ be the source sentence of length $N$, and $\vec{y}_{1:T}=\{\vec{y}_t\}_{t=1}^T$ be the target sentence of length $T$.
In encoder-decoder models, an encoder is used to obtain deterministic context representations of the source sentence, i.e. the encoder hidden states: 
$\vec{h}_{1:N} = f_{enc}(\vec{x}_{1:N})$, where $f_{enc}(\cdot)$ is a nonlinear transition function implemented by LSTM \citep{sutskever2014sequence} or Transformer \citep{vaswani2017attention}.

To introduce context-aware variations
into the encoder, we propose to learn a stochastic function that maps the deterministic hidden states to variables.
Specifically, after computing the hidden states $\vec{h}_{1:N}$ from the transition function $f_{enc}(\cdot)$, the proposed method employs a stochastic mapping function $g(\cdot)$ to model the deterministic context representations as a series of random context variables: 
\begin{equation}
\label{eq:p_z}
    p(\vec{z}_{1:N} \mid \vec{h}_{1:N}) = g(\vec{h}_{1:N}) + \vec{\epsilon}
\end{equation}
where $\vec{\epsilon}\sim \mathcal{N}(\vec{0}, \sigma^2\mat{I})$ is a Gaussian noise.
Then, the decoder can generate diverse texts conditioning on different sets of context variables sampled from $p(\vec{z}_{1:N} \mid \vec{h}_{1:N})$, as shown in Figure \autoref{fig:seq2seq-gp}.
Considering that natural language texts are always context-dependent, we expect the random context variables $\vec{z}_{1:N}$ to encode the context dependency to some extent.
In other words, the distribution of $\vec{z}_i$ {will} not only depend on $\vec{h}_i$, but also depend on other $\{\vec{z}_{j}\}_{j\not= i}$, as shown in Figure \autoref{fig:vae-gp}.


Under this framework, variational encoder-decoder models \citep{bahuleyan2018variational,deng2018latent,ijcai2019-727} can be viewed as a special case, as illustrated in Figure \autoref{fig:seq2seq-normal}, where random context variables $\{\vec{z}_i\}_{i=1}^N$ are \emph{independent} from each other. 
In this work, we consider this special case as generation with {normal priors}.
Empirical comparison between normal priors and GP priors will be given in \autoref{sec:exp}.

\subsection{Gaussian Process Priors for Stochastic Functions}
\label{subsec:prior}

The learning of stochastic function $g(\vec{h})$ is the key for the proposed method to be successful. 
Intuitively, we design $g(\vec{h})$ to satisfy two constraints simultaneously: (1) it can introduce some variation to the deterministic encoder hidden states; (2) it should preserve the contextual information in the deterministic encoder hidden states to be a faithful representation. 

In this work, we propose to learn $g(\vec{h})$ with a functional prior defined by Gaussian processes.
As shown in Figure \autoref{fig:gp-prior} in \autoref{app:gaussian}, we can sample very different functions $g(\vec{h})$ from the same GP prior, which ensures randomness when sampling $\vec{z}_{1:N}$.\footnote{Please refer to \autoref{app:gaussian} and \citep{rasmussen2006gaussian} for detailed introduction of Gaussian processes.}
We define the stochastic function $g(\vec{h})$ following a GP prior:
\begin{equation}
\label{eq:gp-prior}
  g(\vec{h}) \sim \mathcal{GP}(m(\vec{h}), k(\vec{h}, \vec{h}'))
\end{equation}
with the mean function $m(\vec{h})$ and covariance function $k(\vec{h},\vec{h}')$ as 
\begin{equation}
  \begin{split}
    \label{eq:gp-ours}
    m(\vec{h}) &= \vec{h} \\
    k(\vec{h}, \vec{h}') &= 
    v^2 \exp\{-\frac{\|\vec{h}-\vec{h}'\|_2^2}{2r^2}\} 
  \end{split}
\end{equation}
where $\vec{h}$ indicates the current observed encoder hidden state, and $\vec{h}'$ indicates the other contextual encoder hidden states; 
$v$ controls the average distance between a sampled function $g(\vec{h})$ and the mean function $m(\vec{h})$, and $r$ controls the covariance between random variables, increasing $r$ will make $\vec{z}$ and $\vec{z}'$ become more correlated.
In this work, $v$ and $r$ are chosen based on the text generation performance on development sets.
By setting $m(\vec{h})=\vec{h}$, we actually define a semi-parametric GP prior \citep[Sec. 15.2.6]{murphy2012machine} instead of a fully non-parameteric prior, since $\vec{h}$ as a hidden state is computed from the deterministic encoder with learnable parameters.
The intuition behind this definition is that, although we want to introduce some variations, taking the expectation of the sampled random states $\vec{z}$ should still be $\vec{h}$. 

The {main advantage}
of applying GP priors is that we can sample infinite number of random functions $g(\vec{h})$ thus obtaining infinite sets of random context variables $\vec{z}_{1:N}$, as illustrated in Figure \autoref{fig:seq2seq-gp}.
In contrast, standard variational encoder-decoder models can only learn a fixed set of $C$ joint distributions $p(\vec{z}_{1:N}|\vec{h}_{1:N})$, where $1\le C \ll \infty$.\footnote{When $C=1$, it represents a conventional variational autoencoder \citep{bowman-etal-2016-generating}; when $C=5$, it represents a variational autoencoder with a mixture of Gaussians prior (component number = 5); when $C\to \infty$, it represents a variational autoencoder with a GP prior.}

\subsection{Generation with Random Context Variables}
\label{subsec:generation}


In this section, we demonstrate how to incorporate $\vec{z}_{1:N}$ into two typical encoder-decoder models for text generation: a LSTM-based encoder-decoder model \cite{see-etal-2017-get} and a Transformer-based encoder-decoder model \citep{DBLP:journals/corr/abs-1910-10683}.
The performance of these two variational encoder-decoder models with will be evaluated in \autoref{sec:exp}.

Given the deterministic encoder hidden states $\vec{h}_{1:N}$, we first sample a function $g(\vec{h})$ from the GP prior in \autoref{eq:gp-prior}; then sample a set of random context variables $\vec{z}_{1:N}$ from $g(\vec{h})$; finally generate a output sentence $\vec{y}_{1:T}$ based on the sampled $\vec{z}_{1:N}$.
The generative story with random context variables $\vec{z}_{1:N}$ is detailed in \autoref{alg:generative}.

For a LSTM-based encoder-decoder model \cite{see-etal-2017-get}, we apply the attention mechanism \citep{bahdanau2014neural} over the random context variables $\{\vec{z}_i\}_{i=1}^N$ to construct $\vec{c}_{t}$ for the decoder.
At each decoding time step $t$, the decoder computes the attention vector $\vec{c}_t$ and decoder hidden state $\vec{s}_t$ as follows:
\begin{eqnarray}
    \alpha_{ti} &=& \frac{\exp{(a(\vec{s}_{t-1}, \vec{z}_i))}}{\sum_{j=1}^N \exp{(a(\vec{s}_{t-1}, \vec{z}_j))}} \\
    \vec{c}_t &=& \sum_{i=1}^N \alpha_{ti}\cdot \vec{z}_i \\
    \vec{s}_t &=& f_{dec}(\vec{s}_{t-1}, \vec{y}_{t-1}, \vec{c}_t)
  \label{eq:py-yz}
\end{eqnarray}
where $a(\vec{s}_{t-1}, \vec{z}_i) = \vec{v}_a^\top \text{tanh}(\mat{W}_a \vec{s}_{t-1} + \mat{U}_a \vec{z}_i)$, $\mat{W}_a$, $\mat{U}_a$ and $\vec{v}_a^\top$ are parameter matrices.
Finally, the decoder outputs a word distribution based on the context representations and previous decoded words at each decoding time step $t$:
\begin{equation}
  p(\vec{y}_t\mid\vec{y}_{t-1}, \vec{z}_{1:N}) = \text{softmax}(\mat{W}_b \cdot \vec{s}_t)
  \label{eq:p_vocab}
\end{equation}
where $\mat{W}_b$ is a parameter matrix. 

\begin{algorithm}[tb]
  \caption{The generative story with a stochastic function $g(\cdot)$ sampled from the GP prior}
  \label{alg:generative}
  \begin{algorithmic}[1]
    \STATE {\bf Input}: A source sentence $\vec{x}_{1:N}$
    \STATE \textbf{Output}: A generated sentence $\vec{y}_{1:T}$
    \STATE \textit{// Encode context}
    \STATE Initialize $\vec{h}_0 \gets \vec{0}$
    \FOR {$i = 1,\dots,N$}
    \STATE Compute $\vec{h}_{i} = f_{enc}(\vec{h}_{i-1}, \vec{x}_{i})$
    \ENDFOR
    \STATE \textit{// Sample random context variables}
    \STATE Draw $g(\vec{h}) \sim \mathcal{GP}(m(\vec{h}), k(\vec{h}, \vec{h}'))$
    \STATE Draw $\vec{z}_{1:N} \sim g(\vec{h}_{1:N}) + \vec{\epsilon}$
    \STATE \textit{// Generate a new sentence}
    \STATE Initialize $\vec{s}_0 \gets \vec{0}$
    \FOR {$t = 1,\dots,T$}
    \STATE Compute $\vec{s}_t = f_{dec}(\vec{s}_{t-1}, \vec{y}_{t-1}, \vec{z}_{1:N})$
    \STATE Draw $\vec{y}_t \sim \text{softmax}(\mat{W}\cdot \vec{s}_t)$
    \ENDFOR
  \end{algorithmic}
\end{algorithm}

For a Transformer-based encoder-decoder model \citep{DBLP:journals/corr/abs-1910-10683}, we take the output of the last layer in the encoder as $\vec{h}_{1:N}$, and feed them into $g(\vec{h})$ to get random context variables $\vec{z}_{1:N}$.
For the decoder, the inputs $\mat{K}$ and $\mat{V}$ are the combination of $\vec{h}_{1:N}$ and $\vec{z}_{1:N}$:
\begin{eqnarray}
\mat{K}^{l} = \mat{V}^{l} &=& \mat{W}_z[\vec{z}_{1:N}; \vec{h}_{1:N}]\\
\mat{A} &=& \text{MultiHead}(\mat{S}^{l-1}, \mat{K}^{l}, \mat{V}^{l}) \\
\mat{B} &=& \text{LayerNorm}(\mat{A} + \mat{S}^{l-1}) \\
\mat{S}^{l} &=& \text{LayerNorm}(\text{FFN}(\mat{B})+\mat{B})
\end{eqnarray}
where $\mat{S}^{l}=\{\vec{s}_t\}_{t=1}^T$ is the last layer of decoder hidden states, and $\text{MultiHead}(\cdot)$, $\text{LayerNorm}(\cdot)$ and $\text{FFN}(\cdot)$ follow the standard implementation in \citep{vaswani2017attention}.
The word distribution $p(\vec{y}_t\mid\vec{y}_{t-1}, \vec{z}_{1:N})$ at each decoding time step $t$ is computed the same way as in \autoref{eq:p_vocab}.
\section{Efficient Variational Inference}
\label{sec:infer}

With the observed deterministic hidden states $\vec{h}_{1:N}$, 
we estimate the GP posterior to make the prediction of context variables $\vec{z}_{1:N}$ more accurate. 
Although the posterior estimation of Gaussian processes can be written in a closed form theoretically, the challenge in this work comes from learning with other parts of the model, such as the deterministic encoder producing $\vec{h}_{1:N}$ and the decoder generating $\vec{y}_{1:T}$.
To simplify the inference procedure, we will focus on inferring the samples of the GP posterior regarding the hidden states \emph{only} as $p(g \mid \vec{h}_{1:N})$, which essentially is a Gaussian distribution with non-isotropic covariance.
In this work, we apply variational inference to approximate the GP posterior $p(g \mid \vec{h}_{1:N})$ and learn other model parameters \emph{jointly} with maximum likelihood estimation.

For notation simplicity, we let $\vec{h}=f_{enc}(\vec{x}_{1:N})$, $\vec{z}=\{\vec{z}_{i}\}_{i=1}^N$,  $\vec{y}=\{\vec{y}_{i}\}_{i=1}^T$ in this section.
With a {sampled} random function $g(\vec{h})$ from the GP prior as described in line 9 of \autoref{alg:generative}, we will get the joint prior distribution $p(\vec{z}\mid \vec{h})$ according to \autoref{eq:p_z}.
Then we approximate the true posterior $p(\vec{z}\mid \vec{h}, \vec{y})$ with the variational posterior $q_{\vphi}(\vec{z}\mid \vec{h}, \vec{y})$ by maximizing the evidence lower bound of the marginal log-likelihood (ELBo): 
\begin{equation}
  \label{eq:elbo-final}
  \begin{split}
    \log p(\vec{y}\mid \vec{h}) \ge 
    & \mathbb{E}_{q_{\vphi}}[\log p(\vec{y}\mid \vec{z})]\\
    & - \text{KL}[q_{\vphi}(\vec{z}\mid \vec{h},\vec{y})\| p(\vec{z}\mid\vec{h})]
  \end{split}
\end{equation}
where $\vphi$ denotes the variational parameters.
The derivation of \autoref{eq:elbo-final} is presented in \autoref{appd:elbo}.

During generation, we propose a two-step approximation to simplify $q_{\vphi}(\vec{z}\mid \vec{h},\vec{y})$.
First, to maintain the generative property when using the variational distribution, we propose an approximation of the variational distribution $q_{\vphi}(\vec{z}\mid\vec{h},\vec{y})\approx q_{\vphi}(\vec{z}\mid\vec{h})$.
In this case, random context vector $\vec{z}$ will only depend on $\vec{h}$ during inference.
Second, we apply the mean-field amortized variational approximation \citep{kingma2013auto} to approximating
the parameters of $q_{\vphi}(\vec{z}\mid\vec{h})$:
\begin{equation}
\begin{split}
    q_{\vphi}(\vec{z}\mid \vec{h}) &= \prod_{i=1}^N q_{\vphi}(\vec{z}_i\mid \vec{h}_i) \\
    &= \prod_{i=1}^N \mathcal{N}(f_\mu(\vec{h}_i), f_{\sigma^2}(\vec{h}_i))
\end{split}
\end{equation}
where 
$f_\mu(\cdot)$ and $f_{\sigma^2}(\cdot)$ are the mean and covariance in the amortized variational inference network. 
In this work, we use two simple feed-forward neural networks $f_\mu$ and $f_{\sigma^2}$.
The implementation details are included in \autoref{appd:exp-setup}.

\begin{table*}[t]
  \centering
  \small
  \begin{tabular}{p{0.45\textwidth}|p{0.45\textwidth}}
    \toprule
    \multicolumn{2}{l}{\textbf{Paraphrase Generation:} \textit{Twitter URL Paraphrasing Corpus}}\\
    \midrule
    \emph{Original Sentence} & \emph{Target Paraphrase}\\
    [0.3em]
    Amazon only needs a minute of human labor to ship your next package.  &  Amazon ships your packages in one minute.  \\
    Amazon only needs a minute of human labor to ship your next package. &  Amazon only needs a minute of labor to ship your next package. \\
    \midrule
    \midrule
    \multicolumn{2}{l}{\textbf{Text Style Transfer:} \textit{GYAFC Corpus}} \\
    \midrule
    \emph{Informal Sentence} & \emph{Formal Sentence}\\
    [0.3em]
    I’d say it is punk though.  &  However, I do believe it to be punk.  \\
    Gotta see both sides of the story. & You have to consider both sides of the story. \\
    \bottomrule
  \end{tabular}
  \caption{\label{tab:dataset}
    Some example source-target pairs from the Twitter URL Paraphrasing Corpus \cite{lan2017continuously} and the GYAFC Corpus \cite{rao-tetreault-2018-dear}.
  }
\end{table*}

\section{Experiments}
\label{sec:exp}

We evaluate our method on two text generation tasks that require rich contextual representations: paraphrase generation (\autoref{sec:exp-paraphrase}) and text style transfer (\autoref{sec:exp-style}).
We provide some data examples for the two tasks in \autoref{tab:dataset}.
We compared our method with previous works on diverse text generation in terms of quality and diversity.
Empirical experiment results show that our method is able to: 
(1) adapt to different encoder-decoder architectures, such as the pointer-generator network \citep[PG]{see-etal-2017-get} and the text-to-text transfer Transformer \citep[T5]{DBLP:journals/corr/abs-1910-10683};
(2) generate higher quality texts compared with deterministic encoder-decoder models \citep{see-etal-2017-get,DBLP:journals/corr/abs-1910-10683} while also enabling diverse generation by conditioning on random context variables.

\subsection{Evaluation Methods}
\paragraph{Text Quality.}
For quality evaluation, we use two commonly used automatic metrics in text generation: \textbf{METEOR} \cite{banerjee2005meteor} and \textbf{BLEU} with up to bi-grams \cite{papineni2002bleu}, which tell us how well the generated outputs match the reference sentences.

\paragraph{Text Diversity.}
For diversity evaluation, we aim at examining how well different latent context variables $\vec{z}_{1:N}$ from $q_{\vphi}(\vec{z}\mid \vec{h}, \vec{y})$ can make the decoder generate diverse outputs.
We use self-BLEU with up to bi-grams \cite[\textbf{self-BLEU}]{zhu2018texygen} to measure the mutual bi-gram overlap between the set of outputs per source sentence, lower self-BLEU indicates less bi-gram overlap between generated outputs.
In addition, we use diverse 4-gram \cite[\textbf{Div-4}]{deshpande2019fast} to measure the ratio of distinct 4-grams in the set of outputs per source sentence, higher diverse 4-gram shows more unique 4-grams between generated outputs.
Finally, we use uniqueness \cite[\textbf{Uni.}]{deshpande2019fast} to measure the ratio of unique generated sentences in the set of outputs per source sentence, higher uniqueness suggests different context variables $\vec{z}_{1:N}$ lead to very different output sentences.

\subsection{Competitive Baselines}
We compare our method with competitive deterministic encoder-decoder models and variational encoder-decoder models as follows.

\newparagraph{PG}: \citet{see-etal-2017-get} proposes the pointer-generator network, which is a strong deterministic LSTM-based encoder-decoder baseline. We refer their model as {PG}.

\newparagraph{T5}: \citet{DBLP:journals/corr/abs-1910-10683} propose the text-to-text transfer Transformer, which is a strong deterministic Transformer-based encoder-decoder baseline. We refer their model as {T5}.

\newparagraph{Variation Attention}: \citet{deng2018latent} model the deterministic attention vectors as latent alignment variables to promote diverse text generation. We refer their model as {Variation Attention}.

\newparagraph{Multi-Selectors}: \citet{cho-etal-2019-mixture} use a mixture of experts to sample different binary masks on the source texts for diverse content generation. We refer their model as {Multi-Selectors}.

\newparagraph{T-CVAE}: \citet{ijcai2019-727} model deterministic encoder hidden states as latent context variables with a Transformer-based conditional variational autoencoder. We refer their model as {T-CVAE}. 

\newparagraph{PG/T5 + Normal prior}: PG or T5 with a {normal prior} $p(\vec{z})=\mathcal{N}(\vec{0}, \mat{I})$, which follows the conventional variational autoencoders \cite{kingma2013auto,bowman-etal-2016-generating}.

\newparagraph{PG/T5 + GP prior}: PG or T5 with our GP prior defined in \autoref{eq:gp-prior}.

\begin{table*}[t!]
  \centering
  \small
  \begin{tabular}{lcccccc}
    \toprule
    & \multicolumn{2}{c}{\textsc{Twitter URL}} & \multicolumn{2}{c}{\textsc{GYAFC (E\&M)}} & \multicolumn{2}{c}{\textsc{GYAFC (F\&R)}} \\
    \cmidrule(lr){2-3} \cmidrule(lr){4-5} \cmidrule(lr){6-7} 
    Methods & BLEU$\uparrow$ & METEOR$\uparrow$ & BLEU$\uparrow$ & METEOR$\uparrow$ & BLEU$\uparrow$ & METEOR$\uparrow$ \\
    \midrule 
    \emph{Seq2Seq baselines}\\
    PG & 0.291 & 0.471 & {0.683} & 0.817 & 0.717 & 0.845  \\
    T5 & 0.264 & 0.453 & {0.683} & {0.819} & 0.726 & 0.847 \\
    \midrule
    \emph{Related works}\\
    Multi-Selectors & 0.290 & 0.492 & 0.606 & 0.779 & 0.618 & 0.783 \\
    Variation Attention & 0.294 & \textbf{0.512} & 0.632 & 0.804 & 0.675 & 0.833 \\
    T-CVAE & \textbf{0.339} & 0.494 & 0.481 & 0.686 & 0.537 & 0.730 \\
    \midrule
    \emph{Our works}\\
    PG + Normal prior & 0.041 & 0.127 & 0.145 & 0.354 & 0.191 & 0.452 \\
    T5 + Normal prior & 0.269 & 0.461 & 0.675 & 0.815 & 0.722 & 0.846 \\
    PG + GP prior & {0.307} & 0.483 & {0.681} & \textbf{0.828} & {0.734} & \textbf{0.849} \\
    T5 + GP prior & 0.281 & 0.474 & \textbf{0.688} & 0.815 & \textbf{0.739} & 0.847 \\
    \bottomrule
  \end{tabular}
  \caption{\label{tab:pg-acc}
  Model performance on text quality.
  To get the best performance of variational encoder-decoder models, we directly take the mean of  $q_{\phi}(\vec{z}\mid \vec{h}, \vec{y})$ as the sampled context variables to generate the output texts.
  Results of the Multi-Selectors \citep{cho-etal-2019-mixture}, Variation Attention \citep{deng2018latent} and T-CVAE \citet{ijcai2019-727} are collected based on the source code provided in the original paper.
  }
\end{table*}

\subsection{Generation Setups}
For the decoding strategy, we use beam search with beam size of 10. 
Note that our method is orthogonal to all diversity-promoting decoding strategies, such as top-k sampling \citep{fan-etal-2018-hierarchical} and nucleus sampling \citep{holtzman2019curious}.
We choose beam search in order to make fair comparison with other works which promotes diversity at the encoder side.

For quality generation, we directly take the mean of $q_{\vphi}(\vec{z}| \vec{h}, \vec{y})$, and generate one $\vec{y}_{1:T}$ based on the sampled context variables $\vec{z}_{1:N}$, since we want to examine how well the posterior network can encode contextual information and make the decoder generate high-quality texts.

For diverse generation, we sample different $\vec{z}_{1:N}$ (instead of directly taking the mean) from $q_{\vphi}(\vec{z}| \vec{h}, \vec{y})$, and generate different $\vec{y}_{1:T}$ based on the sampled context variables $\vec{z}_{1:N}$, since we want to examine how well different latent context variables $\vec{z}_{1:N}$ from $q_{\vphi}(\vec{z}\mid \vec{h}, \vec{y})$ can make the decoder generate diverse outputs.
For experiment setups, we sample 10 different $\vec{z}_{1:N}$ and generate 10 different $\vec{y}_{1:T}$ correspondingly. 
We compute the diversity scores following the prior work \citep{deshpande2019fast}.
To compute the self-BLEU and Div-4, we randomly sample 5 different $\vec{y}_{1:T}$ out of the 10 generated $\vec{y}_{1:T}$.
To compute the Uni., we compute the unique number of sentences among the 10 generated $\vec{y}_{1:T}$. 

In preliminary experiments, we found that sampling from the original variational distribution tends to make the decoder generate same sentences.
We hypothesize that $q_{\vphi}(\vec{z}| \vec{h}, \vec{y})$ is a high-dimensional multivariate Gaussian and sampling from a high-dimensional distribution is a fundamental challenging problem.
Therefore, we applied a simple heuristics to alleviate the sampling issue, where we scale up the covariance matrix of the variational distribution by a numeric scalar.
We find this simple heuristics can help the decoder generate more diverse sentences.
For PG + Normal prior and PG + GP prior, we set the numeric scalar for paraphrase generation task to 25, for style transfer task to 10.
For T5 + Normal prior and T5 + GP prior, we set the numeric scalar for paraphrase generation task to 7, for style transfer task to 4.

\subsection{Paraphrase Generation}
\label{sec:exp-paraphrase}
We first evaluate the model's capability of generating paraphrases using the Twitter URL paraphrasing dataset \cite{lan2017continuously}.
In this task, we aim at comparing the quality of generated texts between our method and other competitive baselines.
We include the experimental setups in \autoref{appd:exp-setup}.

\paragraph{Dataset.}
The Twitter URL paraphrasing dataset \cite{lan2017continuously} contains both positive and negative examples of paraphrases.
We filter out all negative examples from the 1-year 2,869,657 candidate pairs, and divided the remaining paraphrase pairs into 110K training pairs, 3K testing pairs and 1K validation pairs.

\begin{table*}[t!]
  \centering
  \small
  \begin{tabular}{lcccccccc}
    \toprule
    & \multicolumn{4}{c}{\textsc{GYAFC (E\&M)}} & \multicolumn{4}{c}{\textsc{GYAFC (F\&R)}} \\
    \cmidrule(lr){2-5} \cmidrule(lr){6-9} 
    Methods & avg-BLEU$\uparrow$ & self-BLEU$\downarrow$ & Div-4$\uparrow$ &  Uni.$\uparrow$ & avg-BLEU$\uparrow$ & self-BLEU$\downarrow$ & Div-4$\uparrow$ & Uni.$\uparrow$ \\
    \midrule 
    T-CVAE & 0.481 & 0.986 & 0.221 & 0.130 & 0.537 & 0.990 & 0.220 & 0.126 \\
    T5 + Normal prior & 0.419 & 0.522 & 0.524 & 0.791 & 0.347 & 0.415 & 0.484 & 0.845 \\
    T5 + GP prior & 0.329 & \textbf{0.395} & \textbf{0.727} & \textbf{0.898} & 0.252 & \textbf{0.295} & \textbf{0.748} & \textbf{0.910} \\
    \bottomrule
  \end{tabular}
  \caption{\label{tab:pg-diversity}
  Model performance on text diversity.
  Avg-BLEU measures the average quality of generated sentences compared with ground-truth references. 
  Self-BLEU measures the token-level repetitiveness, and a lower self-BLEU indicates a higher token-level diversity.
  Div-4 measures the ratio of unique 4-grams, and a higher Div-4 means a higher token-level diversity.
  Uni. measures the ratio of unique generated sentences, and a higher Div-4 illustrates a higher sentence-level diversity.
  We include the sampling configuration details in \autoref{appd:exp-setup}.
  }
\end{table*}

\paragraph{Result Analysis.}
As shown in \autoref{tab:pg-acc}, for the quality of generated texts, our method is able to well preserve the semantic information from source texts.
For LSTM-based models, PG + GP prior generates better quality texts compared with both its deterministic baseline PG and other variational baselines, e.g. Multi-Selectors and Variation Attention.
Note that PG + Normal prior experiences the posterior collapse problem \citep{bowman-etal-2016-generating,pmlr-v80-kim18e,pmlr-v89-dieng19a} during training, which causes the context variables preserving little semantic information in the source text and the model generating random tokens during inference.
For Transformer-based models, T-CVAE generates the better quality texts than T5, T5 + Normal prior and T5 + GP prior.
But T-CVAE lowercases all input and output tokens while the other models keep both lowercase and capital tokens, this text preprocessing step may bring an unfairly better performance of T-CVAE in quality scores.
Note that the posterior collapse problem does not happen in T5 + Normal prior, and T5 + GP prior still outperforms T5 + Normal prior, which shows the advantage of GP priors in introducing context-aware variations.

\subsection{Text Style Transfer}
\label{sec:exp-style}
We evaluate our model's capability of generating stylistic texts using the Grammarly's Yahoo Answers Formality Corpus (GYAFC) \cite{rao-tetreault-2018-dear}.
In this task, we first compare the quality of generated texts between our method and other competitive baselines, then we test the diversity of generated texts between our GP prior and conditional variational autoencoders.
We include the experimental setups in \autoref{appd:exp-setup}.

\paragraph{Dataset.}
The GYAFC dataset covers two sub-domains: Entertainment \& Music (E\&M), which has 52,593 training pairs, 2,877 validation pairs, 1,416 testing pairs; and Family \& Relationships (F\&R), which has 51,967 training pairs, 2,788 validation pairs, 1,332 testing pairs.

\begin{table*}[t!]
  \centering
  \small
  \begin{tabular}{p{0.2\textwidth}|p{0.35\textwidth}|p{0.35\textwidth}}
    \toprule
     \multicolumn{3}{p{0.95\textwidth}}{\textit{Informal Sentence}: Your age... Dude that one is old.} \\
     \multicolumn{3}{p{0.95\textwidth}}{\textit{Formal References}: ["Your age. That one is old."; "You are quite old."; "Wow, that one is very old."; "How old are you?"]} \\
    \midrule
    \textbf{T-CVAE} & \textbf{T5 + Normal Prior} & \textbf{T5 + GP Prior} \\
    you are older . & Your age is old. & Your age, that one is old. \\
    you are older . & Your age, that one is old. & Your age, that one is old. \\
    you are older . & You're your age. No, that one is old. & Your age, and that one is old. \\
    you are older . & You are your age. Due, that one is old. & You are your age, and that one is old. \\
    you are older . & Your age, that one is old......... & Your age, you are not the one who is old. \\
    you are older . & Your age and i........ & You're a fool, that one is old. \\
    you are older . & Your age is arbitrary to you......... & Your age doesn't matter, that one is old. \\
    you are older . & You are a very, jo, you are a very, jo & Your age is due to the fact is very old. \\
    you are older . & You are a ant / a ante / a ante / a ante / & Regardless of your age, he is a young person. \\
    you are older . & Your count count count count count count & I am not sure your age, but that one is old. \\
    \bottomrule
  \end{tabular}
  \caption{\label{tab:z-em}
    Sample outputs conditioned on different $\vec{z}$ sampled from $q_\phi(\vec{z}|\vec{x})$ on GYAFC (E\&M) test set.
  }
\end{table*}

\paragraph{Result Analysis.}
For the quality of generated texts, GP prior makes the model more robust to generate accurate texts.
As shown in \autoref{tab:pg-acc},
for LSTM-based models, PG + GP prior generates the most accurate texts compared with PG, Multi-Selectors and Variation Attention. 
Note that PG + Normal prior also experiences the posterior collapse problem in GYAFC datasets, resulting in very low quality scores on the test set.  
For Transformer-based models, T5 + GP prior achieves the best performance than T5, T5 + Normal and T-CVAE, which shows the superiority of GP priors in encoding contextual information.

For the diversity of generated texts, imposing context-aware variations into encoder hidden states is beneficial for generating diverse outputs.
As demonstrated in \autoref{tab:pg-diversity}, 
for Transformer-based models, T5 + GP prior gives the best diversity performance in both token-level and sentence-level compared with T5 + Normal prior and T-CVAE.
The model performance on the style transfer task verifies the capability of our GP prior in promoting generation diversity.
\autoref{tab:z-em} shows some diverse generation outputs of Transformer-based variational encoder-decoder models.
However, we also notice that increasing diversity will inevitably cause degradation in quality, because $\vec{z}_{1:N}$ are i.i.d. sampled from a high-dimensional multivariate Gaussian $q_\phi(\vec{z}\mid \vec{h}, \vec{y})$. 
As discussed in previous work \citep{vono2022high}, multivariate sampling in high-dimensional settings can become computationally demanding.

\paragraph{Computation Complexity Analysis.}
Our GP priors require more computation during training, where the major computation comes from calculating the full co-variance matrix of context variables of the GP prior. 
However, during inference, we approximate the GP posterior with a variational posterior $q_\phi(\vec{z}\mid \vec{h}, \vec{y})$ and conducts i.i.d. sampling, which saves the time for multivariate sampling and has the same computation complexity with other conditional variational autoencoder baselines at testing time.

\section{Related Works}
\label{sec:related}

\paragraph{Diverse text generation.}
Related works on diverse text generation mainly focus on changing decoding strategies at the decoder side or introducing randomness at the encoder side.
At the decoder side, recent works apply various decoding algorithms to promote diversity, such as diverse beam search \cite{vijayakumar2016diverse}
, top-k sampling \cite{fan-etal-2018-hierarchical} 
and nucleus sampling \cite{holtzman2019curious}.
Our model is orthogonal to these diverse decoding algorithms since we focus on the encoder side.
Another group of works \cite{he2018sequence,shen2019mixture} propose to use a mixture of decoders to generate multiple outputs, where the context encodings are shared across multiple decoders.
At the encoder side, \citet{cho-etal-2019-mixture} propose to leverage a mixture of selectors to identify key contents from the source text, where each selector samples a sequential binary latent variables as a hard attention mask on every source token.
\citet{xu2018d} train different pattern embeddings, and generate diverse paraphrases conditioning on different pattern embeddings.

\paragraph{Conditional variational autoencoders.}
Variational encoder-decoder models \cite{deng2018latent,bahuleyan2018variational,ijcai2019-727,sun-etal-2021-generating} are related to our method.
\citet{deng2018latent} formulate the attention vector as latent alignment variables, and use the latent variables as hard attention for the decoder to select which source words to focus on during generation.
\citet{ijcai2019-727} present a conditional variational autoencoder based on Transformer, and learn a latent variable for generating diverse texts for the story completion task.
\citet{sun-etal-2021-generating} propose a self-separated conditional variational autoencoder that introduces group information to regularize the latent variables, which alleviates the posterior collapse problem and improves the model performance in the dialogue generation task.

\section{Conclusion}
\label{sec:con}

In this work, we investigate the problem of generating high quality texts for variational encoder-decoder models.
We propose a novel stochastic function to introduce context-aware variations into encoder hidden states, which provides the decoder with more diverse contextual representations.
To learn this stochastic function, we propose a GP prior to model the dependency between random context variables, and apply an efficient amortized variational inference method to approximate the GP posterior. 
Experimental results demonstrate that our method can learn a better contextual representation that leads to higher generation quality compared with deterministic encoder-decoder models and conditional variational autoencoders.

\bibliography{naaclhlt2019,anthology}
\bibliographystyle{acl_natbib}

\appendix

\section{Gaussian Processes as Function Priors}
\label{app:gaussian}

\begin{figure*}[ht]
\subfigure[Sample $g$ from Gaussian process prior.]{
  \includegraphics[width=0.45\linewidth]{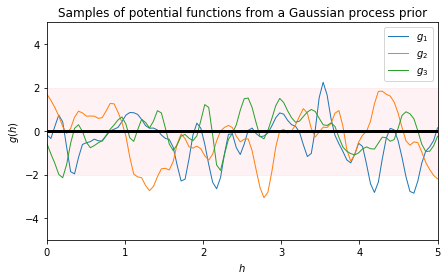}
  \label{fig:gp-prior}
 }
\subfigure[Sample $g$ from Gaussian process posterior.]{
  \includegraphics[width=0.5\linewidth]{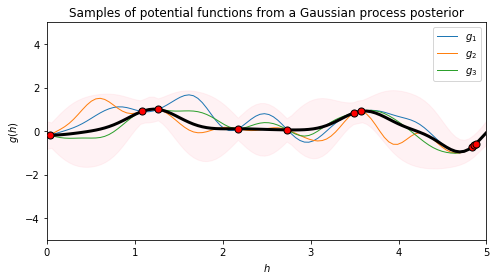}
  \label{fig:gp-post}
}
\caption{Samples of potential functions $g$ from (a) a Gaussian process prior $\mathcal{GP}(\vec{0}, \mat{K})$ which uses a squared exponential kernel $k(h, h')=\exp(-\frac{(h-h')^2}{2})$, (b) a Gaussian process posterior $p(g\mid \mathcal{D})$ conditioning on the training data $\mathcal{D}$.}
\end{figure*}

In this work, we are interested in learning a non-linear mapping function from encoder hidden states to latent context variables.
Gaussian process has the nice property which can represent complex non-linear functions and also allow uncertainty to account for noisy data observations.
Considering a set of observed training data $\mathcal{D} = \{(\vec{h}_i, \vec{z}_i)\}_{i=1}^N$, a Gaussian process defines a probability distribution over possible functions $p(g)$.
Given a Gaussian process prior $\mathcal{GP}(\vec{0}, \mat{K})$ on the function $g(\vec{h})$, we have:
\begin{eqnarray}
    \vec{z}_i &=& g(\vec{h}_i) + \epsilon_i \\
    g(\vec{h}_i) &\sim& \mathcal{GP}(\vec{0}, \mat{K}) \\
    \epsilon_i &\sim& \mathcal{N}(\vec{0}, \sigma^2\mat{I})
\end{eqnarray}
Note that $\epsilon_i$ is the noise of the observed data point $\vec{z}_i$, which is assumed to be an independent identically distributed Gaussian with variance ${\sigma}^2$.
$\mat{K}$ is the covariance matrix which is constructed using a squared exponential covariance function $k(\vec{h}, \vec{h}')=\exp(-\frac{\left\Vert \vec{h}-\vec{h}' \right\Vert^2}{2})$.
Now, we can sample different mapping functions $g(\vec{h})$ from this Gaussian process prior $\mathcal{GP}(\vec{0}, \mat{K})$.
Figure \autoref{fig:gp-prior} illustrates some possible mapping functions $g_1$, $g_2$ and $g_3$.

In Gaussian process, each training and testing data point is treated as random variable which follows Gaussian distribution.
Therefore, we can apply the Bayesian inference to predict a testing data point $\vec{z}_{*}$ conditioning on observed training data points $\mathcal{D}$.
To make concise notation, we let $\vec{h}_{1:N}=\{\vec{h}_i\}_{i=1}^N$, $\vec{z}_{1:N}=\{\vec{z}_i\}_{i=1}^N$, and $\mat{K}^{-1}=[k(\vec{h}_{1:N}, \vec{h}_{1:N})+{\sigma}^2\mat{I}]^{-1}$.
The probability distribution of the testing data point $\vec{z}_{*}$ can be computed by:
\begin{equation}
    p(\vec{z}_{*} \mid \vec{h}_*,  \mathcal{D}) = \int p(\vec{z}_{*} \mid \vec{h}_*, \vec{g}, \mathcal{D})p(\vec{g} \mid \mathcal{D}) dg
\end{equation}
where
\begin{equation}
\begin{split}
\label{eq:exact-post}
    & p(\vec{z}_{*} \mid \vec{h}_*,  \mathcal{D}) \sim \mathcal{N}(\vec{\mu}_*, \mat{K}_*) \\
    & \vec{\mu}_* = k(\vec{h}_{*}, \vec{h}_{1:N})\mat{K}^{-1}\vec{z}\\
    & \mat{K}_* = k(\vec{h}_{*}, \vec{h}_{*}) - k(\vec{h}_{*}, \vec{h}_{1:N})\mat{K}^{-1} k(\vec{h}_{1:N}, \vec{h}_{*})
\end{split}
\end{equation}
Intuitively, training data points $\mathcal{D}$ constrain the set of functions $g$ to pass through them since the covariance becomes smaller when we have training data, as shown in Figure \autoref{fig:gp-post}.

Under our variational encoder-decoder framework, $\vec{h}_{1:N}$ are encoder hidden states and $\vec{z}_{1:N}$ are latent context variables.
Since the Gaussian process induces a distribution over the mapping function $g(\vec{h})$, theoretically we could sample infinite number of mapping functions, where each function gives us a different set of latent context representations $\vec{z}_{1:N}$.
In this way, we managed to obtain diverse context representations in encoder-decoder models.

\section{Derivations of ELBo}
\label{appd:elbo}
We follow conditional variational autoencoders \cite{NIPS2015_8d55a249} and assume that for given observation $\vec{h}$, $\vec{z}$ is drawn from the
prior distribution $p(\vec{z}\mid \vec{h})$, and the output $\vec{y}$ is generated from the distribution $p(\vec{y}\mid \vec{h}, \vec{z})$.
We learn the variational posterior by minimizing $\text{KL}(q_{\vphi}(\vec{z}\mid\vec{h}, \vec{y}) \Vert p(\vec{z}\mid \vec{h}, \vec{y}))$, which is equivalent to maximizing the evidence lower bound of the marginal log-likelihood (ELBo): 
\begin{equation}
\small
\begin{split}
    & \text{KL}(q_{\vphi}(\vec{z}\mid \vec{h}, \vec{y}) \Vert p(\vec{z}\mid \vec{h}, \vec{y})) \\
    &= \int q_{\vphi}(\vec{z}\mid \vec{h}, \vec{y}) \log \frac{q_{\vphi}(\vec{z}\mid \vec{h}, \vec{y})}{p(\vec{z}\mid \vec{h}, \vec{y})} d{\vphi} \\
    &= \int q_{\vphi}(\vec{z}\mid \vec{h}, \vec{y}) \log \frac{q_{\vphi}(\vec{z}\mid \vec{h}, \vec{y})p(\vec{y}\mid \vec{h})p(\vec{h})}{p(\vec{z}, \vec{h}, \vec{y})} d{\vphi} \\
    &= \log p(\vec{y}\mid \vec{h}) \\
    &+ \int q_{\vphi}(\vec{z}\mid \vec{h}, \vec{y}) \log \frac{q_{\vphi}(\vec{z}\mid \vec{h}, \vec{y})p(\vec{h})}{p(\vec{y}\mid \vec{h}, \vec{z})p(\vec{z}\mid \vec{h})} d{\vphi} 
\end{split}
\end{equation}
Since $\text{KL}(q_{\vphi}(\vec{z}\mid \vec{h}, \vec{y}) \Vert p(\vec{z}\mid \vec{h}, \vec{y})) \ge 0$, 
we have:
\begin{equation}
\small
\begin{split}
    & \log p(\vec{y}\mid \vec{h}) \\
    &\ge -\int q_{\vphi}(\vec{z}\mid \vec{h}, \vec{y}) \log \frac{q_{\vphi}(\vec{z}\mid \vec{h}, \vec{y})p(\vec{h})}{p(\vec{y}\mid \vec{h}, \vec{z})p(\vec{z}\mid \vec{h})} d{\vphi} \\
    &= \mathbb{E}_{q_{\vphi}}[\log p(\vec{y}\mid \vec{z}, \vec{h})+\log p(\vec{z}\mid \vec{x}) \\
    &~~~~~~~~~~~~ - q_{\vphi}(\vec{z}\mid \vec{h}, \vec{y})]\\
    &= \mathbb{E}_{q_{\vphi}}[\log p(\vec{y}\mid \vec{z}, \vec{h})] - \mathbb{E}_{q_{\vphi}}[\frac{q_{\vphi}(\vec{z}\mid \vec{h}, \vec{y})}{\log p(\vec{z}\mid \vec{h})}] \\
    &= \mathbb{E}_{q_{\vphi}}[\log p(\vec{y}\mid \vec{z}, \vec{h})] \\
    &~~~~~~~~~~~~ - \text{KL}[q_{\vphi}(\vec{z}\mid \vec{h}, \vec{y}) \Vert p(\vec{z}\mid \vec{h})]
\end{split}
\end{equation}
where $\vphi$ are parameters for the variational inference networks.

\section{Experiment Setup Details}
\label{appd:exp-setup}

\paragraph{Model Configurations.}
For the implementation details of PG, it is an LSTM-based encoder-decoder model with copying mechanism.
The encoder is a 1-layer Bi-LSTM, and the decoder is a 1-layer uni-directional LSTM.
We set the word embedding size to 300, the hidden dimension for both encoder and decoder to 512.
We let the encoder and decoder shares the same vocabulary list and word embedding, and the vocabulary size is 20000.
For the configuration of the posterior networks, both the mean and covariance network are a single feed-forward neural network, and we set the dimension of the latent variable to 256.

For the implementation details of T5, we use the T5-base implementation from Huggingface \citep{wolf-etal-2020-transformers} \footnote{\url{https://huggingface.co/transformers/model_doc/t5.html}}, and use their default model configuration.
We load the pre-trained weights of T5-base, and fine-tune them on our target task datasets.
For the configuration of the posterior networks, both the mean and covariance network are a single feed-forward neural network, and we set the dimension of the latent variable to 512.

\paragraph{Training Configurations.}
For the training details of PG and T5, we does not apply KL annealing and the coefficient of the KL divergence is always 1.
We use Adam optimizer \cite{duchi2011adaptive} with learning rate of 0.0001, and adopt early stopping if the validation loss does not decrease after 10 epochs.
For the hyper-parameters $\{v, r\}$ of the kernel function in \autoref{eq:gp-ours}, we try a range of values where $v \in [0.01, 100]$ and $r \in [0.0001, 10]$, and do grid search cross validation on the validation set to select the best model.  
All experiments are independently conducted on a GPU server (RTX 2090 Ti) with 40cores CPU and 256GB Memory.

\end{document}